\pdfoutput=1
\documentclass{article}
\usepackage{mlspconf,spconf, amsmath,graphicx}
\usepackage[shortlabels]{enumitem}
\usepackage{color}
\usepackage[table]{xcolor} 
\usepackage{ifthen}
\usepackage[colorlinks]{hyperref}
\pdfoutput=1

\usepackage{fancyhdr}

\title{\center{ScatterNET HYBRID DEEP LEARNING (SHDL) NETWORK \\ FOR OBJECT CLASSIFICATION}}
\name{Amarjot Singh, Nick Kingsbury}
\address{Signal Processing Group, Department of Engineering, University of Cambridge, U.K.}

\toappear{2016 IEEE International Workshop on Machine Learning for Signal Processing, Sept.\ 13--16, 2016, Salerno, Italy}

\begin{document}

%
\maketitle
\begin{abstract}
The paper proposes the ScatterNet Hybrid Deep Learning (SHDL) network that extracts invariant and discriminative image representations for object recognition. SHDL framework is constructed with a multi-layer ScatterNet front-end, an unsupervised learning middle, and a supervised learning back-end module. Each layer of the SHDL network is automatically designed as an explicit optimization problem leading to an optimal deep learning architecture with improved computational performance as compared to the more usual deep network architectures. SHDL network produces the state-of-the-art classification performance against unsupervised and semi-supervised learning (GANs) on two image datasets. Advantages of the SHDL network over supervised methods (NIN, VGG) are also demonstrated with experiments performed on training datasets of reduced size.
\end{abstract}
\section{Introduction}
\label{sec:intro}

Object classification is challenging due to the large intra-class variability, arising from translation and rotation of objects, lighting, deformations and occlusions. Researchers have relied on invariant and discriminative class-specific image representations to tackle this problem~\cite{Kavukcuoglu}. 

Numerous attempts have been made to design learning architectures that capture the necessary image representations for object classification. These methods include architectures that: (i) encode handcrafted features extracted from the input images into rich \textit{non-hierarchical} representations~\cite{ponce}; (ii) learn multiple levels of \textit{feature hierarchies}, directly from the input data~\cite{Kavukcuoglu}; (iii) make use of the ideas from both the above-mentioned categories to extract \textit{feature hierarchies} from \textit{hand-crafted features}. 

Bag of Words (BoW)~\cite{google} models represent the first class of architectures that encode handcrafted bag-of-visual-words (BOV) descriptors into rich feature representations using unsupervised coding and pooling~\cite{ponce}. The pipeline was improved by encoding the local descriptor patches by a set of visual codewords with sparse coding with a linear SPM kernel~\cite{yang}. This class of methods are very easy to design and cheap to evaluate but achieve only marginally good classification performance on different benchmarks~\cite{yang}.

The second category includes architectures such as Convolutional Neural Networks (CNNs)~\cite{NIN,vgg}, Deep Belief Networks~\cite{DBN} etc that learn feature hierarchies directly from the input images. These networks have achieved state-of-the-art classification performance on various datasets~\cite{NIN,vgg}, but despite the success of these networks, their design and optimal configuration is not well understood which makes it difficult to develop them. In addition, these models produce a large number of coefficients that are learned with the help of powerful computational resources and require large training datasets which may not be available for many applications such as stock market prediction~\cite{su}, medical imaging~\cite{medical} etc.

The third class of models combines the concepts from both the above-mentioned models to learn feature hierarchies from low-level hand-crafted descriptors~\cite{sivic}. Hierarchical max (HMAX)~\cite{serre} is one such model that uses an RBF kernel to learn a single layer of high-level features from descriptors captured with a battery of Gabor filters.  He et al.~\cite{Kavukcuoglu} learned three layers of sparse hierarchical features from SIFT descriptors using unsupervised learning. Sivic et al.~\cite{sivic}, discovered object class hierarchies from visual codewords using hierarchical LDA. This class of models has produced promising performance on various datasets~\cite{nomp,lowe}. In addition, each layer of these models can be posed as an optimization problem resulting in optimal architectures~\cite{Kavukcuoglu}.

The paper introduces the ScatterNet Hybrid Deep Learning (SHDL) network for object classification. This framework first extracts ScatterNet handcrafted descriptors which are used by an unsupervised learning module to learn hierarchical features that capture intricate structure between different object classes. Supervised learning then selects the features specific to each object class, from the feature hierarchies, which are finally used for classification. The term 'Hybrid' is coined because the framework uses both unsupervised as well as supervised learning. Each layer of the network is designed and optimized automatically that produces the desired computationally efficient architectures.

 \begin{figure*}[t!] 
\centering    
\includegraphics[width = 0.99\textwidth, height = 10.0 cm]{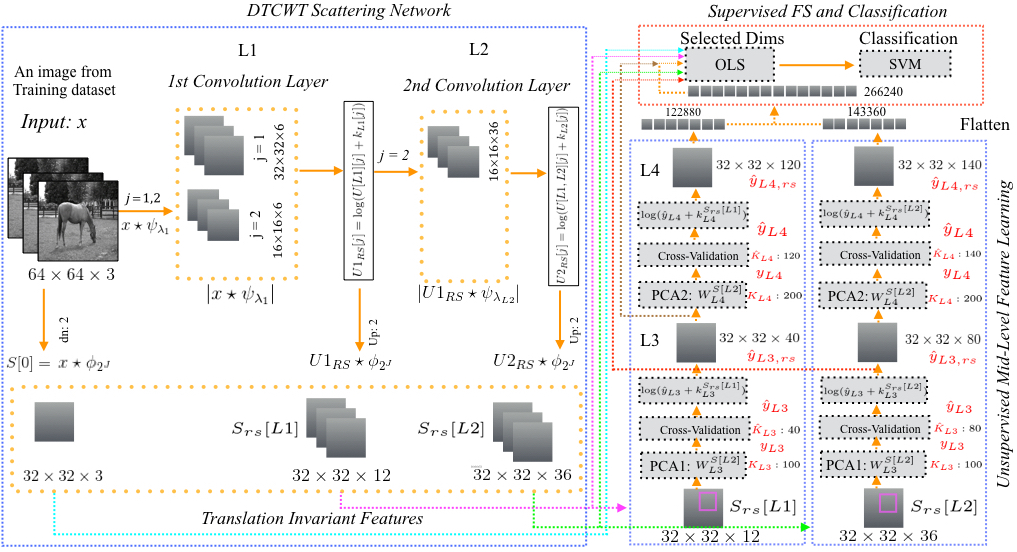}
\caption[Scattering operations performed on a signal to extract filtered response with translation invariance and to recover the lost frequency components. ]{\small{SHDL: The illustration shows the input image ($64 \times 64$ ($x$)) from the CIFAR-10 dataset at resolution R1 decomposed to extract the translation invariant relatively symmetric coefficients at L0 ($S_{rs}[L0]$), L1 ($S_{rs}[L1]$) and L2 ($S_{rs}[L2]$). Features at the higher level of abstraction are captured at L3 and L4 layers of the PCA-Net using unsupervised learning. Parametric log transformation is applied on the output of each PCA stage to introduce relative symmetry. The representations extracts at each stage (L0, L1, L2, L3, L4) are concatenated and given to the supervised OLS layer that select the object-specific features finally used for classification using the Gaussian SVM (G-SVM).}}
\label{fig:scatter00}
\end{figure*} 
  
The contributions of the paper are as follows:

\begin{itemize}[topsep=2.2pt]
\itemsep0em 
\vspace{-0.4em}
\item \textit{Hand-crafted Module}: The hand-crafted descriptors are extracted with the two-layer parametric log ScatterNet~\cite{singh}, instead of BOW~\cite{google} or SIFT~\cite{lowe} descriptors. This extracts symmetrically distributed multiscale oriented edge features at the first layer and additional discriminative sparse features at the second layer~\cite{Jbruna2013}. 

\vspace{-0.2em}
\item \textit{Unsupervised Learning Module}: This module uses two stacked PCA-Net~\cite{pcanet} layers with parametric log non-linearity to learn robust symmetrically distributed hierarchical mid-level features across object classes. The network is fast to train as opposed to other unsupervised learning modules (autoencoders or RBMs) as the minimization of the loss function (Eq. 8) can be obtained in its simplistic form as the Eigen decomposition. 
\vspace{-0.2em}	
\item \textit{Supervised Learning Module}: OLS layer~\cite{Blumensath,singh} is applied to the concatenated features obtained from the layers to select a subset of object-class-specific features, without undesired bias from outliers, due to the introduced symmetry. The selected features are fed into a Gaussian-kernel support vector machine (G-SVM) to perform object classification. 

\item \textit{Network Layer Optimization}: The number of filters in each layer of the unsupervised learning module are optimized as part of the automated design process. The optimization of the number of filters in a layer leads to the efficient learning of the subsequent layer as the filters are now learned from a smaller feature space. The reduced feature space subsequently also makes the learning of the OLS and SVM efficient. 
  \end{itemize}

The classification performance of the proposed architecture is tested on CIFAR-10 and Caltech-101 datasets. Multiple experiments on different training dataset sizes are performed to highlight the advantages of the proposed network against supervised and unsupervised methods.

Section 2 of the paper briefly presents the proposed SHDL network. Section 3 presents the experimental results while Section 4 draws conclusions.
\vspace{-0.7em}
\section{SHDL Network}
\label{sec:typestyle}
This section introduces the proposed ScatterNet Hybrid Deep Learning (SHDL) network (Fig. 1) with the detailed mathematical formulation of each module and a description of the representations captured by them. 
\vspace{-0.7em}
\subsection{Hand-crafted Module: ScatterNet}
Handcrafted descriptors are extracted using the parametric log based two-level Dual-Tree Complex Wavelet Transform (DTCWT) ScatterNet~\cite{singh} that extracts relatively symmetric translation invariant low-level features at the first layer and more discriminative sparse features at second layer~\cite{Jbruna2013,ima,eccv}. The extracted features are also dense over the scale as they are obtained by decomposing multi-resolution images obtained at 1.5 times (R1) and twice (R2) the size of the input image. This ScatterNet~\cite{singh} is chosen over Bruna and Mallat's~\cite{Jbruna2013} due to its superior classification accuracy and computational efficiency. The DTCWT ScatterNet formulation is presented for an input signal ($x$) which may then be applied to each multi-resolution image.

\begin{figure}[t!] 
\centering    
\includegraphics[width=1\linewidth, height = 4.28 cm]{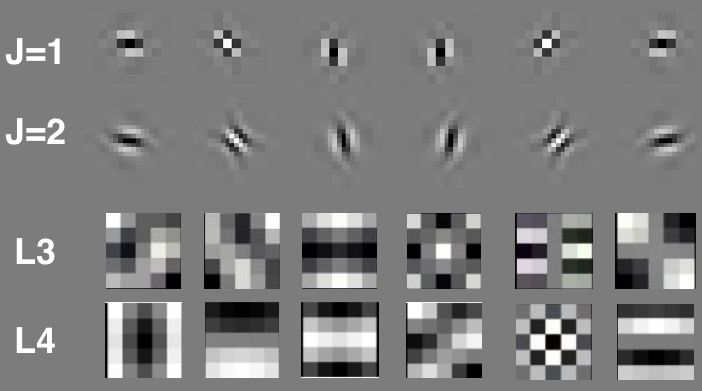}
\caption[Scattering operations performed on a signal to extract filtered response with translation invariance and to recover the lost frequency components. ]{\small{Illustration shows the DTCWT real filters at two scales used at Layer L1 and L2. The filters learned by the PCA-Net at L3 and L4 stage are also shown.}}
\label{fig:scatter00}
\end{figure}

The features at the first layer are obtained by filtering the input signal $x$ with dual-tree complex wavelets $ \psi_{j,r }$ at different scales ($j$) and six pre-defined orientations ($r$) fixed to $15^\circ, 45^\circ, 75^\circ, 105^\circ, 135^\circ$ and $165^\circ$, as shown in Fig. 2. A more translation invariant representation is built by applying a point-wise $L_{2}$ non-linearity (complex modulus) to the real and imaginary part of the filtered signal:
\vspace{-0.2em}
\begin{equation}
U[{L1}] =  \sqrt{|x\star \psi_{\lambda_{1} }^{a}|^2 + |x\star \psi_{\lambda_{1} }^{b}|^2} 
\end{equation}
The parametric log transformation layer is applied on the oriented features, extracted at the first scale $j=1$ with a parameter $k_{L1}[j]$, to reduce the effect of outliers by introducing relative symmetry (rs) to their amplitude distribution (as shown in Fig. 3): 
 \vspace{-0.2em}  
   \begin{equation}
  U1_{rs}[j] = \log(U[L1][j] + k_{L{1}}[j]), \quad U[L1][j] = |x\star \psi_{j}|, 
\end{equation}
The parameter $k_{L{1}}$ is selected such that it minimizes the difference between the mean and median of the distribution~\cite{singh}. Next, a local average is computed on the envelope $|U1[\lambda_{L1}]|$ that aggregates the coefficients to generate the desired translation-invariant representation: 
\vspace{-0.3em}
\begin{equation}
S_{rs}[L1] = |U1_{rs}| \star \phi_{2^J}
\end{equation}
The energy (high-frequency components) lost due to smoothing is recovered by cascaded wavelet filtering applied at the second layer~\cite{Jbruna2013}. The recovered components are again not translation invariant so invariance is achieved by first applying the $L_{2}$ non-linearity to obtain the regular envelope:
\vspace{-0.2em}
\begin{equation}
U[L2] = |U1_{rs} \star \psi_{\lambda_{L2}}|
\end{equation}
The parametric log transformation is applied again to produce relative symmetry: 
 \vspace{-0.3em}  
   \begin{equation}
  U2_{rs}[j] = \log(U[L2][j] + k_{L2}[j])
\end{equation}
Next, a local-smoothing operator is applied to improve translation invariance:: 
\vspace{-0.3em}
\begin{equation}
S_{rs}[L2] = U2_{rs}\star \phi_{2^J}
\end{equation}
The output coefficients are typically formed from $x\star\phi$ (Layer 0), $S_{rs}[L1]$ (Layer 1) and $S_{rs}[L2]$ (Layer 2) for each of the two image resolutions R1 and R2.

\begin{figure}[t!] 
\centering    
\includegraphics[width=1\linewidth, height = 1.9 cm]{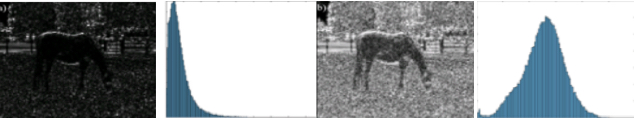}
\caption[Scattering operations performed on a signal to extract filtered response with translation invariance and to recover the lost frequency components. ]{\small{Illustration shows a representation obtained using the DTCWT ($15^\circ$, j=1, R=1) at L1. The representation is affected by outliers resulting in the skewed distribution. A relatively symmetric representation is obtained using by applying the parametric log transformation that also results in contrast normalization.}}
\label{fig:scatter00}
\end{figure}
\vspace{-0.96em}
\subsection{Unsupervised Learning Module: PCA-Net Layers}
\label{PCA}
This section details the optimization framework for the stacked PCA-Net~\cite{pcanet} layers used to learn symmetrically distributed hierarchical mid-level features at $L3$ and $L4$, from invariant features extracted at $L1$ or $L2$, as shown in Fig. 1. The mathematics is presented for $S_{rs}[R1,L1]$ (invariant features obtained for R1 resolution image at layer L1). This formulation is also applied to $S_{rs}[R1,L2]$ (features for R1 resolution at layer L2) as well as features extracted at R2 resolution at both layers ($S_{rs}[R2,L1]$ and $S_{rs}[R2,L2]$).

The objective of the PCA layer is to minimize the reconstruction error by learning a family of multi-channel orthonormal filters. In order to learn the filters, $M$ overlapping patches of size $z_1\times z_2$ are collected from each channel of the input $S[R1,L1]$ i.e., $ x_{1}, x_{2},..., x_{M} \in {R}^{z_1z_2 \times P} $ where $x$ is the sampled patch, $M$ represents the number of patches and $P$ (12 and 36, Fig. 1) represents the number of channels of the input. After this, the patch mean is subtracted to obtain $\tilde{ X} = [\tilde{  x}_{1},\tilde{ x}_{2},...,\tilde{ x}_{M}]$, where $\tilde{ x}$ is a mean-removed patch. Given $N$ training images, we get the unified matrix: 
   \vspace{-0.3em}
\begin{equation}\label{eq: datamatrix_1}
 X = [\tilde{ X}_1,\tilde{ X}_2,...,\tilde{ X}_N]\in {R}^{z_1z_2M \times PN}.
\end{equation}
This filters are learned by minimizing the following equation, 
\vspace{-0.3em}
\begin{equation}
\min_{ W_{L3}\in R^{s{L_{3}}s{L_{3}} \times P \times K_{L3}}} \| X -  W_{L3} W_{L3}^T\ X\|_F^2,~{\rm s.t.}~  W_{L3}^T W_{L3} =   I_{K_{L3}},
\end{equation} 
where $W_{L3}$ are the learned filters at layer $L_{3}$ with size $s{L_{3}} \times s{L_{3}} \times P \times K_{L3}$, where $K_{L3}$ represents the number of filters. The solution in its simplified form represents $K_{L3}$ principal eigenvectors of $ X X^T$. These learned filters (Fig. 2) capture the variance in the training dataset in the form of eigenvectors. 

The output responses of the $L3$ layer can be obtained as:
\vspace{-0.3em}
\begin{equation}
y_{L3} = S_{rs}[R1,L1]\star W_{L3}^{s{L_{3}} \times s{L_{3}} \times P \times K_{L3}}   , i = 1,2,3,.....,N
\end{equation}
$S_{rs}[R1,L1]$ is zero-padded before convolving with $W_{L3}$ so as to make $y_{L3}$ have the same size as $S_{rs}[R1,L1]$.

 Next, G-SVM is used at the output of the L3 layer, with varying number of filters (10,20,..,$K_{L3}$), to select the optimal number ($\hat{K}_{L3}$) of learned filters that result in the highest five-fold cross-validation accuracy (5-CV) on the training dataset (Fig. 3). The optimum output a L3 layer ($\hat{y}_{L3}$) is computed using Eq. 10 using the optimum number of filters ($\hat{K}_{L3}$)

As explained in the previous section, parametric log transformation is applied on $\hat{y}_{L3}$ to introduce relative symmetry: 
 \vspace{-0.3em}  
   \begin{equation}
  \hat{y}_{L3,rs} = \log(\hat{y}_{L3}+ k_{L3})
\end{equation}

\begin{figure}[t!] 
\centering    
\includegraphics[width=1\linewidth, height = 3.7 cm]{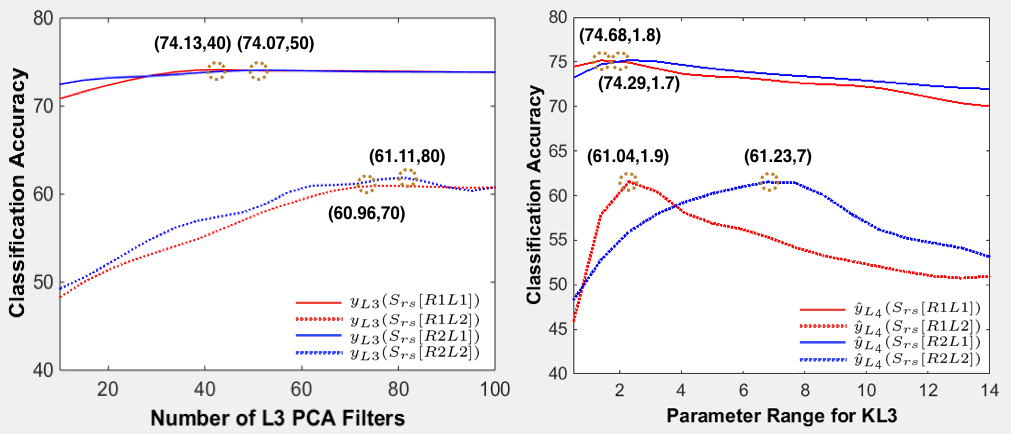}
\caption[Scattering operations performed on a signal to extract filtered response with translation invariance and to recover the lost frequency components. ]{\small{Illustration shows: (a) 5-CV classification accuracy ($y_{L3}$) vs. the number of filters ($K_{L3}$) learned at layer L3 (b) optimal 5-CV classification accuracy ($\hat y_{L3}$) vs. $k_{L3}$, with fixed number of optimal L3 filters ($\hat K_{L3}$). The optimal filters $\hat K_{L3}$ and chosen $k_{L3}$ along with their corresponding accuracies is shown in the graphs.}}
\label{fig:scatter00}
\end{figure} 
  
Next, $K_{L4}$ filters with weights $W_{4}$ at layer $L4$ can be learned similarly:
\vspace{-0.3em}
\begin{equation}
\begin{aligned}
\min_{ W_{L4}\in R^{s{L_{4}}s{L_{4}} \times K_{3} \times K_{L4}}} \| X^{L3} -  W_{L4} W_{L4}^T\ X^{L3}\|_F^2,~{\rm s.t.} \\ W_{L4}^T W_{L4} =  I_{K_{L4}},
\end{aligned}
\end{equation} 

where $X^{L3}$ represents the matrix computed by extracting patches from $\hat{y}_{L3,rs}$ (\textit{L3 output (relatively symmetric (rs)) obtained using the optimal ($\hat{K}_{L3}$) number of filters}). The output response at Layer $L4$ can be computed as shown:
\vspace{-0.3em}
\begin{equation}
y_{L4} = \hat{y}_{L3,rs}\star W_{L4}^{s{L_{4}} \times s{L_{4}} \times K_{L3} \times K_{L4}}   , i = 1,2,3,.....,N
\end{equation}
Here, $\hat{y}_{L3,rs}$ is also zero padded before applying the convolutions as described above. The optimal L4 output ($\hat{y}_{L4}$) is computed using $\hat{K}_{L4}$ filters, obtained using five-fold cross-validation as shown in Fig. 4.  Parametric log transformation is finally applied on $\hat{y}_{L4}$ to introduce relative symmetry: 
 \vspace{-0.3em}  
   \begin{equation}
  \hat{y}_{L4,rs} = \log(\hat{y}_{L4}+ k_{L4})
\end{equation}
\vspace{-1.4em}
\subsection{Supervised Learning Module: OLS and G-SVM}
The features obtained from each layer of the network (L0, L1, L2, L3, L4) for both R1 and R2 images are concatenated, normalized across each dimension and fed to the OLS as shown in Fig. 1. Orthogonal least square (OLS) regression~\cite{Blumensath}selects discriminative features specific to class $C$ in a supervised way using a one-versus-all linear regression. The regression is applied to the training set of scattering features where each vector of $N$ (Cifar: N $\approx$ 176000, Caltech: N $\approx$ 474000) dimensions is reduced to $N'$ (Cifar: N' $\approx$ 10300, Caltech: N' $\approx$ 21000) selected dimensions. The reduced training feature dataset is utilized by the G-SVM to learn weights that best discriminate the classes in the dataset. Feature selection results in limited dimensions that lead to a more efficient training of the G-SVM and improves generalization. 
\begin{figure}[t!] 
\centering    
\includegraphics[width=1\linewidth, height = 3.7 cm]{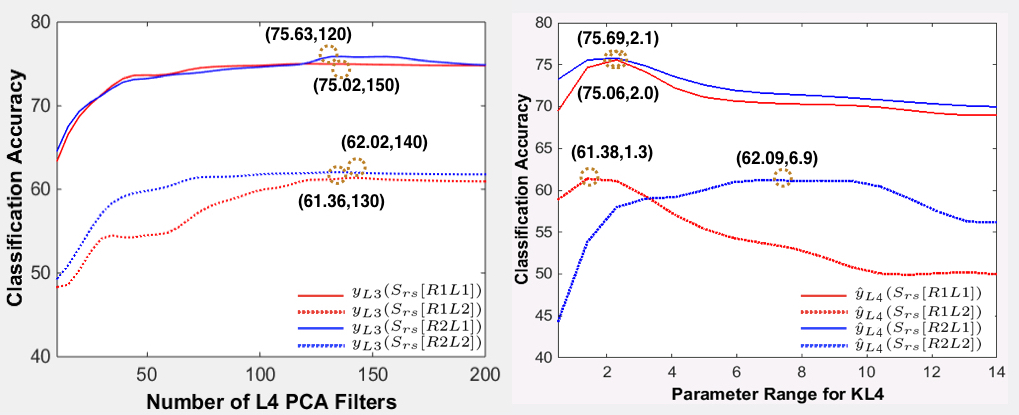}
\caption[Scattering operations performed on a signal to extract filtered response with translation invariance and to recover the lost frequency components. ]{\small{Illustration shows: (a) 5-CV classification accuracy ($y_{L4}$) vs. the number of filters ($K_{L4}$) learned at layer L4 (b) optimal 5-CV classification accuracy ($\hat y_{L3}$) vs. $k_{L4}$, with fixed number of optimal L4 filters ($\hat K_{L4}$). The optimal filters $\hat K_{L3}$ and chosen $k_{L3}$ along with their corresponding accuracies is shown in the graphs.}}
\label{fig:scatter00}
\end{figure} 
\vspace{-0.4em}
\section{Overview of Results}
\label{headings}
The performance of the SHDL network is evaluated on CIFAR-10 and Caltech-101 datasets. CIFAR-10 contains a total of 50000 training and 10000 test images each of size $32\times32$. The Caltech-101 dataset is an unbalanced image dataset with images of different sizes. In these experiments, 30 images (resized to $128 \times 128$) per class (clutter class removed) are used for training, 10 for validation and the rest of the images in each class are used for testing. Average per class classification results are reported with an averaging over 5 random splits. A detailed comparison with unsupervised, semi-supervised, and supervised methods is also presented. 

\subsection{ScatterNet feature extraction}
The scattering representations are extracted by first obtaining multi-resolution images of size (64 $\times$ 64 (R1) and 48 $\times$ 48 (R2)) for CIFAR-10 and (256 $\times$ 256 (R1) and 192 $\times$ 192 (R2)) for Caltech-101, as described in Section. 2.1. The images in the CIFAR dataset are decomposed for each colour channel separately using DTCWT filters at 5 (for R1) and 4 (for R2) scales respectively, while the images in the Caltech dataset are decomposed with at 6 and 5 scales for R1 and R2 resolutions respectively. Next, log transformations are applied to the representations obtained (except at the coarsest scale) for both the R1 and R2 pipeline with parameters $k_{j=1}$ = 1.1, $k_{j=2}$ = 3.8, $k_{j=3}$ = 3.8, $k_{j=4}$ = 7 and $k_{j=4}$ = 6.8 (selected as described in Section. 2.1), obtained by averaging the individual $k$ value for the particular scale for all the images in the training dataset. The classification accuracies for each layer (L0, L1, L2) and the concatenated features (HC = S[L0, L1, L2]) are presented for both resolutions, using G-SVM in Table. 1. L2 features give a less good performance on their own than L1, probably due to their lower energies, but still, give a useful improvement when combined with L1.
\vspace{-1.2em}
\begin{table}[!h]
\centering
\caption{\small{Accuracy (\%) on CIFAR-10 for features extracted at different layers and resolutions. $S_{rs}[Layer]$, HC = S[L0,L1,L2]}}
\label{components}
\begin{tabular}{c|cc|cc|cc}
\hline
 & \small{S[L1]} & \small{$S_{rs}[L1]$} & \small{S[L2]} & \small{$S_{rs}[L2]$} & \small{$HC$} & \small{$HC_{rs}$}\\ \hline 
 {R1} & 71.48 & \cellcolor{gray!50}\textbf{72.58} & 60.34 & \cellcolor{gray!50}\textbf{60.51} & 80.7 & \cellcolor{gray!50}\textbf{81.7} \\ \hline
 {R2} & 72.04 & \cellcolor{gray!50}\textbf{73.39} & 60.12 & \cellcolor{gray!50}\textbf{60.39} & 80.9 & \cellcolor{gray!50}\textbf{81.9} \\
\end{tabular}
\end{table}

\vspace{-1.7em}
\subsection{PCA Layers: features and layer optimization}
The L3 PCA layer of the network is trained on $S_{rs}[R1L1]$, $S_{rs}[R1L2]$, $S_{rs}[R2L1]$ and $S_{rs}[R2L2]$, to learn $K_{L3}$=100 filters of size $s_{L3}$=5. 
Cross-validation is used (as explained in Section. 2.2) on the L3 layer output (${y}_{L3}$) to select the 40, 70, 50 and 80 optimal filters ($\hat{K}_{L3}$) for the four cases, as shown in Fig. 3(a). Next, the relatively symmetric L3 output ($\hat{y}_{L3,rs}$) is obtained by applying a log transformation with $k_{L3}$ = 1.8, 1.9, 1.7 and 7.0 on $\hat{y}_{L3}$, respectively (Fig. 3(b)). 

L4 PCA layer is trained on L3 layer outputs ($\hat{y}_{L3,rs}$) correspondingly to learn 200 ($K_{L4}$) filters, of size 5 ($s_{L4}$). Similarly, 150, 140, 120 and 130 optimal filters ($\hat{K}_{L4}$) are selected for the four cases ($S_{rs}[R1L1]$, $S_{rs}[R1L2]$, $S_{rs}[R2L1]$ and $S_{rs}[R2L2]$), as shown in Fig. 4(a). Next, the relatively symmetric L4 outputs ($\hat{y}_{L4,rs}$) are obtained by applying a log transformation with $k_{L4}$ = 2.0, 1.3, 2.1 and 7.2 on $\hat{y}_{L4,rs}$, respectively, for the four cases, as shown in Fig. 4(b). 

The five-fold cross-validation (5-CV) classification accuracies on CIFAR-10, obtained using G-SVM, at different stages of Layers L3 and L4 are presented in Table 2. There are fewer optimal filters in ($K^{op}_{L3}$, $K^{op}_{L4}$) than the originally learned filters ($K_{L3}$, $K_{L4}$) but produce an equal or higher cross-validation accuracy. This suggests that some of the filters learn redundant information which can be removed. This results in efficient learning of L4 layer (subsequently for OLS as well as SVM) as the L4 filters are learned from a smaller feature space $\hat{y}_{L3,rs}$ (obtained with $\hat{K}_{L3} << {K}_{L3}$). 
\vspace{-0.6em}
\subsection{Classification performance}
This section evaluates the classification performance of each module of the SHDL network. The classification accuracy of each module is presented by applying the supervised OLS layer on the features to select the relevant features which are then fed to the G-SVM to compute the accuracy. The accuracy of the handcrafted module (HC) is computed on the concatenated relatively symmetric features extracted at L0, L1, L2, for both resolutions (R1, R2) using OLS for feature selection and then G-SVM for classification. The hand-crafted module produced a classification accuracy of 82.4\% (HC) on CIFAR-10 as shown in Table. 3. An increase of 0.4\% is observed when the mid-level features, learned at L3 with $s_{L3}$=5 are concatenated with the features of the hand-crafted module (HC,$(L3)_{s_{L3}=5}$), again for both R1 and R2. A further increase of 0.7\% (HC,$(L3,L4)_{s_{L3,L4}=5}$) is noticed when mid-level features from the L4 layer learned with $s_{L4}$=5 are concatenated to (HC,$(L3)_{s_{L3}=5}$) features. This suggests that the PCA layers (L3 and L4) learn useful image representations as they improve the classification performance. Finally, in order to test the optimality of the filter sizes, the L3 and L4 layers were also trained with $s_{L3}$=3 and $s_{L4}$=3. A further increase of around 0.4\% ($HC,(L3,L4)_{3,5}$) is observed by concatenating the features obtained at L3 and L4 layers, with filters trained with the kernel $s_{L3},s_{L4}$ of size 3 and 5, with the hand-crafted module (HC). This suggests that filters of different sizes learn unique and useful image representations. 
\vspace{-0.8em}
\begin{table}[!t]
\centering
\caption{\small{5-CV Accuracy (\%) on CIFAR-10 at L3 and L4. ${y}_{L3}$, ${y}_{L4}$  output, $\hat{y}_{L3}$, $\hat {y}_{L4}$ optimal output and $\hat{y}_{L3,rs}$, ${y}_{L4,rs}$ relatively symmetric output, at L3 and L4.}}
\label{components}
\begin{tabular}{c|ccc|ccc}
\hline
 & $\small{{y}_{L3}}$ & $\small{\hat{y}_{L3}} $ & $\small{\hat{y}_{L3,rs}}$ & $\small{{y}_{L4}}$ & $\small{\hat{y}_{L4}}$ & $\small{\hat{y}_{L4,rs}}$  \\ \hline 
 \small{$S_{rs}[R1L1]$} & \small{73.83} & \small{74.13} & \small{74.68}  & \small{74.81} & \small{75.02} & \small{75.06}  \\
 \small{$S_{rs}[R1L2]$} & \small{60.78} & \small{60.96} & \small{61.04} & \small{60.93} & \small{61.36} & \small{61.38}  \\ \hline
 \small{$S_{rs}[R2L1]$} & \small{73.86} & \small{74.07} & \small{74.29} & \small{74.88} & \small{75.63} & \small{75.69}   \\
 \small{$S_{rs}[R2L2]$} & \small{60.81} & \small{61.11} & \small{61.23} & \small{61.78} & \small{62.02} & \small{62.67}  \\
\end{tabular}
\end{table}

\vspace{-0.3em}
\begin{table}[!h]
\centering
\caption{\small{Accuracy (\%) on CIFAR-10 for each module computed with OLS and G-SVM. The increase in accuracy with the addition of each layer is also shown. HC: Hand-crafted, PCA features ($(Layer)_{filter-size}$): eg $(L3)_{s_{L3}=5}$}}
\label{MNIST}
\begin{tabular}{c|cccc}
\hline
 & \small{HC} & \small{$HC,(L3)_{5}$} & \small{$HC,(L3,L4)_{5}$} & \small{$HC,(L3,L4)_{3,5}$} \\
 \hline
\small{Acc.} & 82.4 & 82.8 & 83.5 & 83.9\\
\end{tabular}
\end{table}
\vspace{-0.3em}

Next, the performance of the SHDL network is evaluated on the Caltech-101 dataset. The network results in a classification accuracy of 81.46\%, as shown in Table. 4. 
\vspace{-0.9em}
\subsection{Comparison with the state-of-the-art}
The SHDL outperformed the semi-supervised and unsupervised learning methods on both datasets however the network underperformed by nearly 13\% against supervised deep learning models~\cite{NIN,vgg}, as shown in Table. 4.  
\vspace{-0.8em}
\begin{table}[!h]
\centering
\caption{\small{Accuracy (\%) and comparison on both datasets. Unsup: Unsupervised, Semi: Semi-supervised and Sup: Supervised.}}
\label{MNIST}
\begin{tabular}{c|ccc|cc}
\hline
 \small{Dataset} & \small{SHDL} & \small{Semi} & \small{Unsup}  & \small{Sup}\\
 \hline
 \small{CIFAR-10} & \cellcolor{gray!50}\textbf{\small{83.90}} & \small{83.3~\cite{GAN}} &\small{82.9~\cite{nomp}}  & \small{96.2~\cite{WRN}}\\
\small{Caltech-101} & \cellcolor{gray!50}\textbf{\small{81.46}} & \small{81.5~\cite{cal-semi}} & \small{81.0}~\cite{lowe}  & \small{92.7~\cite{vgg}}\\

\end{tabular}
\end{table}

\subsection{Advantage over supervised learning}
Supervised models require large training datasets to learn which may not exist for most application. Table. 4 shows that SHDL network outperformed VGG~\cite{vgg} and Network in Network (NIN)~\cite{NIN} on the CIFAR-10 datasets with less than 2k images. The experiments were performed by dividing the training dataset of 50000 images into 8 datasets of different sizes. The images for each dataset are obtained randomly from the full 50000 training dataset. It is made sure that an equal number of images per object class are sampled from the training dataset. The full test set of 10000 images is used for all the experiments. Deeper models like NIN~\cite{NIN} and VGG~\cite{vgg} result in low classification accuracy due to their inability to train on the small training dataset.
\vspace{-1.2em}
\begin{table}[!h]
\centering
\caption{\small{Comparison of SHDL network on accuracy (\%) with two supervised learning methods (VGG~\cite{vgg} and NIN~\cite{NIN} against different training dataset sizes on CIFAR-10.}}
\label{MNIST}

\begin{tabular}{c|ccccccc}
\hline
\small{Arch.}  & \small{500} & \small{1K} & \small{2K} & \small{5K} & \small{10K} & \small{20K} & \small{50K}\\
 \hline
\small{SHDL}  & \cellcolor{gray!50}\small{\textbf{50.3}} & \cellcolor{gray!50}\small{\textbf{57.9}} & \cellcolor{gray!50}\textbf{\small{63.4}} & \small{{68.6}} & \small{{72.3}} & \small{{78.4}} & \small{83.9}\\

\small{NIN} & \small{15.6} & \small{54.5} & \small{61.1} & \cellcolor{gray!50}\textbf{\small{{72.9}}} & \cellcolor{gray!50}\textbf{\small{81.2}} & \cellcolor{gray!50}\textbf{\small{86.7}} & {\small{89.6}}\\

\small{VGG}   & \small{10.3} & \small{10.7} & \small{43.4} & \small{63.4} & \small{72.0} & \small{83.1} & \cellcolor{gray!50}\textbf{\small{92.7}}   \\ 
\end{tabular}
\end{table}
\vspace{-1.8em}
\section{Conclusion}
The paper proposes the SHDL network that uses PCA-Net based unsupervised learning module to learn mid-level features while OLS based supervised learning is used to select features that aid the discriminative SVM learning. It is shown that a very simple PCA based network can learn useful features that can greatly improve the classification performance. The network has also shown to outperform unsupervised and semi-supervised learning methods while evidence of the advantage of SHL network over supervised learning (CNNs) methods is presented for small training datasets.  
\vspace{-0.7em}

\bibliographystyle{IEEEbib}
\bibliography{refs}

\begin{thebibliography}{10}

\bibitem{Kavukcuoglu}
Y.~He et~al.,
\newblock ``Unsupervised feature learning by deep sparse coding,''
\newblock {\em Proceedings of the SDM}, 2014.

\bibitem{ponce}
S.~Lazebnik, C.~Schmid, and J.~Ponce,
\newblock ``Beyond bags of features: Spatial pyramid matching for recognizing
  natural scene categories,''
\newblock {\em IEEE CVPR}, 2006.

\bibitem{google}
J.~Sivic and A.~Zisserman,
\newblock ``Video google: A text retrieval approach to object matching in
  videos,''
\newblock {\em IEEE CVPR}, 2003.

\bibitem{yang}
J.~Yang et~al.,
\newblock ``Linear spatial pyramid matching using sparse coding for image
  classification,''
\newblock {\em CVPR}, 2009.

\bibitem{NIN}
M.~Lin, Q.~Chen, and S.~Yan,
\newblock ``Network in network,''
\newblock {\em arXiv:1312.4400}, 2013.

\bibitem{vgg}
K.~Simonyan and A.~Zisserman,
\newblock ``Very deep convolutional networks for large-scale image
  recognition,''
\newblock {\em ICLR}, 2015.

\bibitem{DBN}
H.~Lee~R. Grosse, R.~Rananth, and A.~Ng,
\newblock ``Convolutional deep belief networks for scalable unsupervised
  learning of hierarchical representation,''
\newblock {\em ICML}, 2009.

\bibitem{su}
S.~Jain et~al.,
\newblock ``A novel method to improve model fitting for stock market
  prediction,''
\newblock {\em International Journal of Research in Business and Technology},
  2013.

\bibitem{medical}
Antonello Pasini,
\newblock ``Artificial neural networks for small dataset analysis,''
\newblock {\em Journal of Thoracic Disease}, vol. 7, no. 11, pp. 2278--2324,
  2015.

\bibitem{sivic}
J.~Sivic et~al.,
\newblock ``Unsupervised discovery of visual object class hierarchies,''
\newblock {\em IEEE CVPR}, 2008.

\bibitem{serre}
T.~Serre et~al.,
\newblock ``Robust object recognition with cortex-like mechanisms,''
\newblock {\em IEEE PAMI}, 2007.

\bibitem{nomp}
T.H.~Lin et~al.,
\newblock ``Stable and efficient representation learning with non negativity
  constraints,''
\newblock {\em ICML}, 2014.

\bibitem{lowe}
S.~McCann and D.~Lowe,
\newblock ``Spatially local coding for object recognition,''
\newblock {\em ACCV}, 2012.

\bibitem{singh}
A.~Singh and N.G. Kingsbury,
\newblock ``Dual-tree wavelet scattering network with parametric log
  transformation for object classification,''
\newblock {\em ICASSP}, 2017.

\bibitem{Jbruna2013}
J.~Bruna and S.~Mallat,
\newblock ``Invariant scattering convolution networks,''
\newblock {\em IEEE PAMI}, vol. 35, pp. 1872 --1886, 2013.

\bibitem{pcanet}
TH~Chan et~al.,
\newblock ``Pcanet: A simple deep learning baseline for image
  classification?,''
\newblock {\em ArXiv:1404.3606}, 2014.

\bibitem{Blumensath}
T.~Blumensath and M.~E. Davies,
\newblock ``On the difference between orthogonal matching pursuit and
  orthogonal least squares,''
\newblock 2007.

\bibitem{ima}
A~Singh et~al.,
\newblock ``Multi-resolution dual-tree wavelet scattering network for signal
  classification,''
\newblock {\em International Conference on Mathematics in Signal Processing},
  2016.

\bibitem{eccv}
S~Nadella et~al.,
\newblock ``Aerial scene understanding using deep wavelet scattering network
  and conditional random field,''
\newblock {\em European Conference on Computer Vision}, 2016.

\bibitem{GAN}
T.~Salimans et~al.,
\newblock ``Improved techniques for training gans,''
\newblock {\em ArXiv:1606.03498}, 2016.

\bibitem{WRN}
S.~Zagoruyko and N.~Komodakis,
\newblock ``Wide residual networks,''
\newblock {\em arXiv:1605.07146}, 2016.

\bibitem{cal-semi}
Dengxin~Dai et~al.,
\newblock ``Unsupervised high-level feature learning by ensemble projection for
  semi-supervised image classification and image clustering,''
\newblock {\em arXiv}, 2016.

\end{thebibliography}

\end{document}